\documentclass[letterpaper, 10 pt, journal, twoside]{IEEEtran} 

\IEEEoverridecommandlockouts 

\usepackage{amssymb}  
\usepackage{amsmath,amsfonts}
\usepackage{algorithmic}
\usepackage{array}
\usepackage[caption=false,font=normalsize,labelfont=sf,textfont=sf]{subfig}
\usepackage{textcomp}
\usepackage{stfloats}
\usepackage{url}
\usepackage{verbatim}
\usepackage{graphicx}
\def\BibTeX{{\rm B\kern-.05em{\sc i\kern-.025em b}\kern-.08em
    T\kern-.1667em\lower.7ex\hbox{E}\kern-.125emX}}
\usepackage{balance}

\usepackage{orcidlink}

\begin{document}

\title{A Palm-Shape Variable-Stiffness Gripper based on 3D-Printed Fabric Jamming}

\author{Yuchen Zhao~\orcidlink{0000-0002-9779-4577} and Yifan Wang~\orcidlink{0000-0003-2284-520X}%
\thanks{This work was supported in part by the Singapore Agency for Science, in part by the Technology and Research (A*STAR) under its MTC IRG award M21K2c0118, in part by the A*STAR AME YIRG award A2084c0162, and in part by the Nanyang Technological University NAP award 020482.} 
\thanks{The authors are with School of Mechanical and Aerospace Engineering, Nanyang Technological University, 50 Nanyang Avenue, Singapore, 639798 
        (email
        {\tt\footnotesize yuchen.zhao@ntu.edu.sg; yifan.wang@ntu.edu.sg})}
\thanks{Digital Object Identifier (DOI): 10.1109/LRA.2023.3266667.}
\thanks{\textsuperscript{\textcopyright} 20XX IEEE.  Personal use of this material is permitted.  Permission from IEEE must be obtained for all other uses, in any current or future media, including reprinting/republishing this material for advertising or promotional purposes, creating new collective works, for resale or redistribution to servers or lists, or reuse of any copyrighted component of this work in other works.}
}

\markboth{IEEE Robotics and Automation Letters. Preprint Version. Accepted April, 2023}
{Zhao \MakeLowercase{\textit{et al.}}: A Palm-Shape Variable-Stiffness Gripper based on 3D-Printed Fabric Jamming} 

\maketitle
\begin{abstract}
Soft grippers have excellent adaptability for a variety of objects and tasks. Jamming-based variable stiffness materials can further increase soft grippers' gripping force and capacity. Previous universal grippers enabled by granular jamming have shown great capability of handling objects with various shapes and weight. However, they require a large pushing force on the object during gripping, which is not suitable for very soft or free-hanging objects. In this paper, we create a novel palm-shape anthropomorphic variable-stiffness gripper enabled by jamming of 3D printed fabrics. This gripper is conformable and gentle to objects with different shapes, requires little pushing force, and increases gripping strength only when necessary.
We present the design, fabrication and performance of this gripper and tested its conformability and gripping capacity.
Our design utilizes soft pneumatic actuators to drive two wide palms to enclose objects, thanks to the excellent conformability of the structured fabrics.
While the pinch force is low, the palm can significantly increase stiffness to lift heavy objects with a maximum gripping force of $17\,$N and grip-to-pinch force ratio of $42$.
We also explore different variable-stiffness materials in the gripper, including sheets for layer jamming, to compare their performances.
We conduct gripping tests on standard objects and daily items to show the great capacity of our gripper design.
\end{abstract}



\section{Introduction}
\IEEEPARstart{R}{ecent} advances in soft robotics enable challenging tasks to be accomplished by robots made of soft and compliant materials~\cite{rus_2015_design,manti_2016_stiffening,hughes_2016_soft,shintake_2018_soft}, such as soft pneumatic actuators~\cite{suzumori_1991_development,ilievski_2011_soft, murray_2015_poroelastic,yang_2016_buckling} and universal granular jamming grippers~\cite{brown_2010_universal,amend_2012_positive}.
Variable-stiffness materials (VSM) further improve the gripping capability~\cite{manti_2016_stiffening,yang_2018_principles}.
For example, shape-memory alloys and polymers (SMA/SMP)~\cite{firouzeh_2015_soft,cianchetti_2014_bioinspired}, low-melting point alloys (LMPA)~\cite{shintake_2015_variable,wang_2017_shape}, jamming-based variable-stiffness materials (JVSM)~\cite{fitzgerald_2020_review,manti_2016_stiffening}, \textit{etc}., have been explored to increase soft gripper's capacity.
Among these, JVSM relies on the jamming transition of discrete elements (granules, fibers or sheets) to increase the material's apparent stiffness under confining pressure.
Comparing to SMA/SMP and LMPA, JVSM typically has a short response time ($\lesssim 1\,$sec), easy pneumatic activation, and can hold the stiffened state without continuous energy consumption~\cite{fitzgerald_2020_review,manti_2016_stiffening}.

To take the full advantage of jamming in soft robotic grippers, both structural design and jamming material matter.
The universal jamming gripper developed by Brown \textit{et~al.} \;~\cite{brown_2010_universal} have discrete granular medium that can flow smoothly to conform to various object shapes in the unjammed state; and the gripper can jam to lift the objects via geometric interlocking, frictional contacts and vacuum suction mechanisms.
Although the conformability of universal jamming gripper is desirable, it needs to push down on the object before picking up, and this pushing force may flatten soft items and can sometimes be even larger than the gripping force~\cite{kapadia_2012_design,licht_2018_partially}.
Another common design is to augment soft anthropomorphic grippers with JVSM directly.
Recent works have paired soft pneumatic actuators with granular jamming~\cite{jiang_2019_variable,li_2017_passive,wei_2016_novel,mitsuda_2021_active}, layer jamming~\cite{gao_2020_novel,zhu_2019_fully}, and hybrid jamming~\cite{yang_2019_hybrid}.
Although the finger design effectively reduces the pushing force needed in the universal gripper design, it loses the conformability the latter offered in the sense that there are less contact area under the fingers compared to the universal gripper design.
Grippers using actuated origami ``magic-ball''~\cite{li_2019_vacuum-driven}, rigid~\cite{noauthor_2021_flex-clamp} or soft robotic surface are also proposed~\cite{aksoy_2020_reconfigurable,cai_2021_pneumatic,xiao_2022_modeling}.
Comparing to separated fingers, the wide surface can more securely grip the object, as the former may lose the object through the gaps between fingers or via finger twisting~\cite{wei_2016_novel}.
Integrating the surface with VSM, which have not been proposed before, has the potential to improve the gripping strength and conformability better than adding VSM to the bending actuators alone.

\begin{figure}[!htbp]
    \centering
    \framebox{\parbox{3.2in}{\includegraphics[width=3.2in]{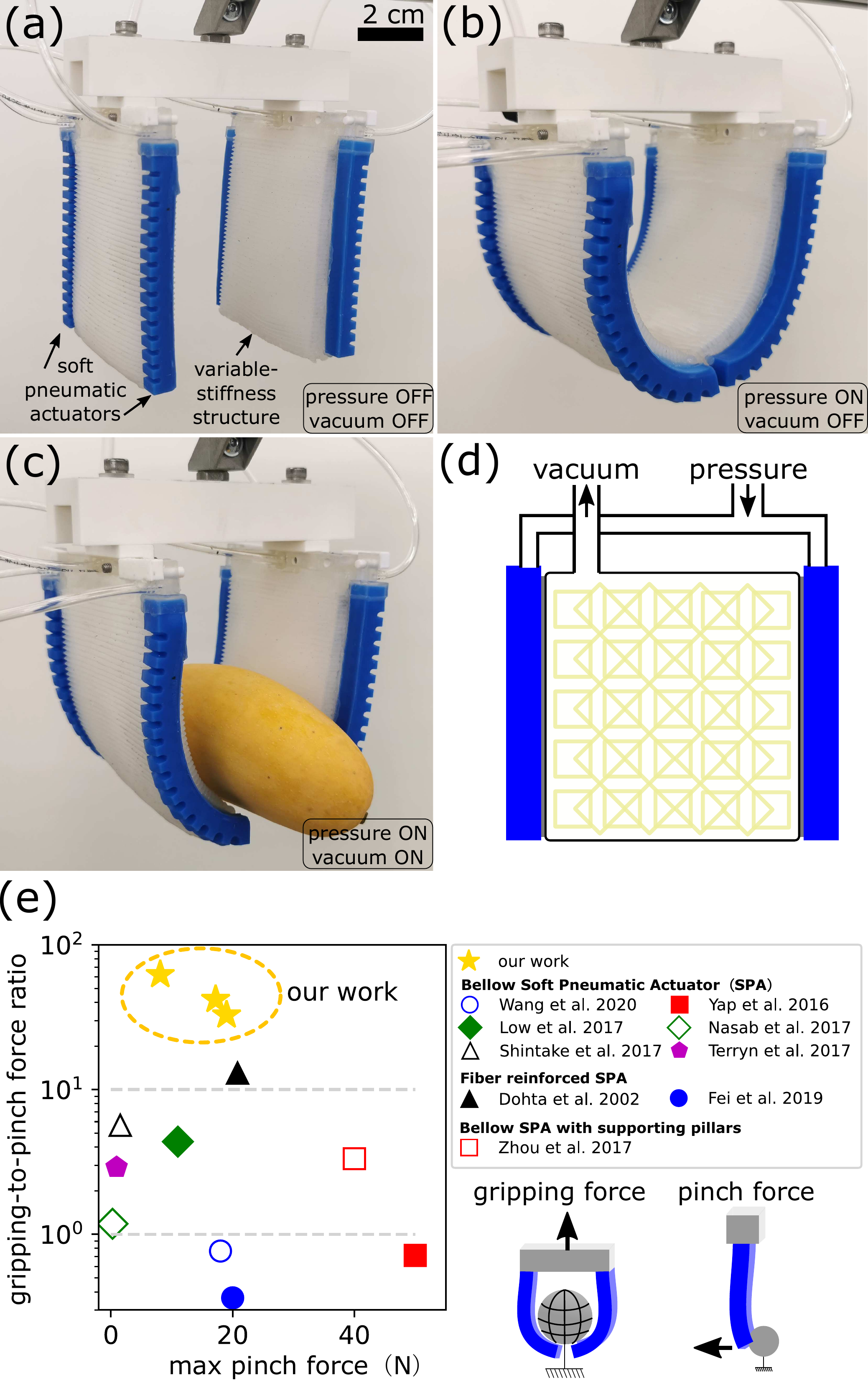}}}
    \caption{Our proposed variable-stiffness gripper is shown in (a). The pneumatic actuators are in blue and the variable-stiffness structures are in white; (b) the palms are bent when the actuators are pressurized to $120\,$kPa; (c) the palms are jammed to hold a mango ($307\,$g); (d) a schematics shows the independent pneumatic wiring for the actuators and variable-stiffness material;
    (e) the ratio of maximum gripping force-to-pinch force ratio is plotted as a function of the maximum gripping force from previous research~\cite{dohta_2002_development,yap_2016_high-force,zhou_2017_soft-robotic,low_2017_hybrid,nasab_2017_soft,shintake_2017_soft,terryn_2017_self-healing,fei_2019_novel,wang_2020_soft}.
    Our work is highlighted in yellow stars.
    The schematics illustrate the orientation of the gripping and pinch force.   
    }
    \label{fig:gripperlook}
\end{figure}

In this paper, we present a palm-shape soft pneumatic gripper equipped with a variable stiffness robotic surface that is achieved by jamming of structured fabrics~\cite{wang_2021_structured} and covers a large portion of the gripper's entire area ($\approx 83\%$).
The gripper actively applies small forces for pinching and can pick up heavy objects via stiffening.
It also has great conformability and does not push down on objects before gripping (i.e. zero pushing force).
Illustrated in Fig.~\ref{fig:gripperlook}a, the gripper consists of two palms, which bend to conform and grip an object (Fig.~\ref{fig:gripperlook}b).
The palms can jam to increase holding stability and hence increase the gripping force (Fig.~\ref{fig:gripperlook}c).
The discrete nature of the fabrics enables our gripper to achieve high conformability as granular jamming-based grippers, and the interlocking structure within fabrics can lead to greater bending stiffness compared to convex granular particles when jammed~\cite{wang_2021_structured}.
Our gripper can reach a maximum gripping force of more than $17\,$N and pick up heavy objects such as a pitaya ($454\,$g), better than previous reported soft robotic surface gripper without variable stiffness materials~\cite{xiao_2022_modeling}.
It can reach a range of maximum gripping force-to-pinch force ratio from $32$ to $62$, higher than previous soft pneumatic actuator gripper equipped with jamming-based variable stiffness materials~\cite{Goh_2022_3d,shintake_2018_soft}(Fig.~\ref{fig:gripperlook}e)
These benefits make our gripper more versatile and adaptable to various gripping tasks in unstructured conditions, such as in packaging, agriculture and food processing, where harvesting and picking non-rigid, free-hanging soft products occurs~\cite{chua_2003_robotic}.

Contributions of our work are summarized as below:
\begin{itemize}
\item[(1)] We incorporate the jamming of structured fabrics in variable-stiffness soft gripper design.
The structured fabric as the jamming medium is lightweight, offer high bending stiffness when jammed, and can be 3D printed with high flexibility in material choice and structure design for specific needs.
\item[(2)] The design offers good conformability and has a high gripping-to-pinching force ratio (Fig.~\ref{fig:gripperlook}e).
\item[(3)] We systematically characterize our grippers through gripping force tests, pinch force measurements, and  conformability test. The results are compared with variable-stiffness grippers based on layer jamming mechanism.
\end{itemize}

The rest of the paper is organized as follows:
We describe the design and fabrication of 3D-printed structured fabrics and soft pneumatic actuators in Sec.~\ref{sec:design and fabrication}.
In Sec.~\ref{sec:characterization}, we use experiments to evaluate the performance of the grippers including object gripping performance, maximum gripping force, pinch force, and gripper conformability.
We also analyze gripping mechanisms based on the experimental results.
We conclude and discuss our work in Sec.~\ref{sec:conclusion}.

\section{Gripper Design and Fabrication}\label{sec:design and fabrication}
\subsection{Variable-Stiffness Palm}
\begin{figure*}[!htbp]
    \centering
    \framebox{\parbox{7.0in}{\includegraphics[width=7.0in]{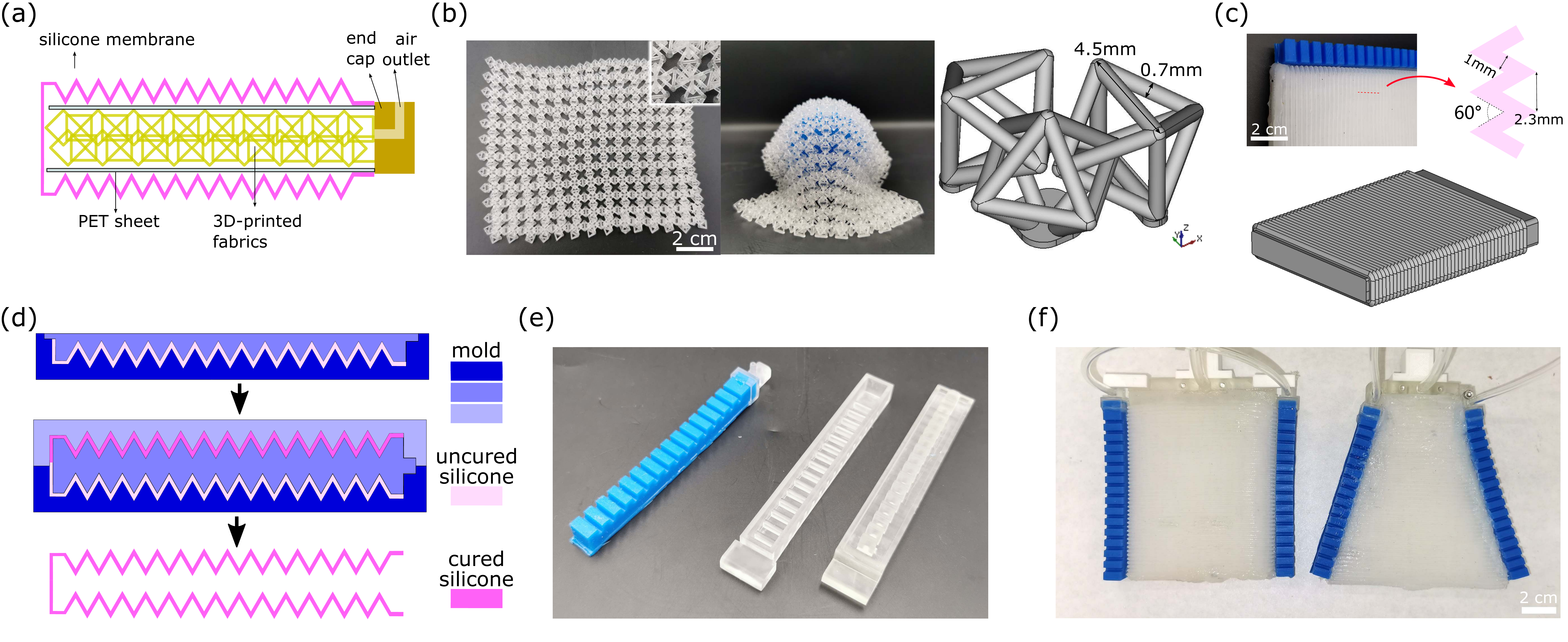}}}
    \caption{Design and fabrication of gripper's main components.
    (a) cross-sectional view of the variable-stiffness composite: silicone membrane (magenta), PET sheets (light-blue with dark boundary), 3D-printed fabrics (yellow), end cap (brown). Drawings are not to scale;
    (b) Left: 3D-printed structured fabrics. The inset is an enlarged picture on the fabric; Middle: the fabric is wrapped onto a $5\,$cm sphere, showing its great conformability; Right: illustration of a unit cell of the fabric;
    (c) Top: silicone membrane and the geometric of its triangular wavy pattern; Bottom: illustration of the complete membrane;
    (d) schematics of the three-step fabrication process of the silicone membrane in cross-sectional view. Blue colors in different shade represent disconnected solid molds. Pink/magenta represents uncured/cured silicone;
    (e) picture of the soft pneumatic actuator (blue) and its mold (semi-transparent);
    (f) the assembled rectangular (left) and trapezoidal (right) palm.}
    \label{fig:fabrication}
\end{figure*}

Fig.~\ref{fig:fabrication}a illustrates the design and fabrication of the gripper's palm components, including structured fabrics, air-tight silicone elastomer membrane, and plastic sheets.
The media used for jamming is based on our previous work on pneumatic jamming of structured fabrics~\cite{wang_2021_structured}.
The fabrics are interlocked regular octahedron trusses, resembling a chain mail structure~(Fig.~\ref{fig:fabrication}b).
It has been shown that the interlocking between the non-convex granular particles can enhance the stiffness variation compared to conventional convex particles~\cite{wang_2021_structured}.
It is manufactured by a stereolithography 3D printer (Formlabs Form3) with rigid material (clear V4 resin).
The octahedron truss diameter and length are $0.7\,$mm and $4.5\,$mm, respectively.
The bottom of each octahedron truss is truncated and attached to a small pad as support, which facilitates 3D printing process and can avoid silicone membrane puncture. 
Two pieces of the fabrics are used in one palm with the padded side faces towards the silicone membrane (Fig.~\ref{fig:fabrication}a).
The fabrics occupy a rectangular volume of about $10 \times 8 \times 1\,\text{cm}^{3}$ in the silicone membrane (made of Smooth-On Ecoflex 30), and the membrane has a size of $10 \times 9.4 \,\text{cm}^{2}$

The membrane has a triangular, zig-zag pattern in the longitudinal direction, as shown in Fig.~\ref{fig:fabrication}c.
The zig-zag pattern alternates at an angle of $60^{\circ}\,$, with a membrane thickness of $1\,$mm and a pitch length of $2.3\,$mm.
The membrane is fabricated via a typical multi-step silicone molding process, as illustrated in Fig.~\ref{fig:fabrication}d.
We first cure one-half of the membrane using 3D-printed molds.
The other half is cured in an extended mold in which the first-half can be placed (Fig.~\ref{fig:fabrication}d middle row).
The membrane's silicone elastomers are cured at room temperature for $4$ hours.
Trade-off exists when using a single membrane material to achieve high or low bending stiffness in the jammed or unjammed state, respectively.
Therefore, composite membrane design or jamming fillers may be employed, such as the sandwich jamming structure investigated for layer jamming~\cite{narang_2020_lightweight}.
We use zig-zag pattern on the membrane surface to reduce the palms' bending stiffness for better conformability in the unjammed state.
We also insert a piece of $0.1\,$mm thin plastic sheet (polyethylene terephthalate) between the fabric and the membrane (Fig.~\ref{fig:fabrication}a).
The plastic sheet is in full contact with the fabric in the jammed state, resulting in a higher bending stiffness than using silicone membrane alone.
Finally, the fabric, membrane and the plastic sheets are all glued to a piece of rigid connector with embedded air outlet to allow them forming a palm.

\subsection{Soft Pneumatic Actuator}
We separately fabricate soft pneumatic actuators to drive the variable-stiffness palm.
The soft pneumatic actuators are based on fast pneumatic network design~\cite{mosadegh_2014_pneumatic}.
Shown in Fig.~\ref{fig:fabrication}e, the actuator is of the same length as the variable-stiffness palm, and has a square cross-section of side length $1\,$cm.
The actuator body is made of silicone elastomer (Smooth-On Mold Star 30).
The strain-limiting layer is a piece of paper or nylon fabric sandwiched between two silicone elastomer sheets (the same material as the body) with a resulting thickness of $1\,$mm.
The silicone elastomer used here is cured at room temperature for $6$ hours.
The strain limiting layer is glued to the air chamber using uncured silicone elastomer mixture.
We use a rigid end cap to seal the actuator, which also serves as an adapter to be connected to the palm.

\subsection{Assembly}
The finished variable-stiffness palm and actuators are connected in parallel using custom fixtures and glues (Fig.~\ref{fig:fabrication}f).
The actuators are glued to the middle plane position of the palm to minimize asymmetric deformation when it is jammed.
Each finished palm has a weight of about $120\,$g, where the 3D-printed fabrics weigh only $22\,$g.
Two such palms are vertically connected to a horizontal beam, forming a parallel jaw-shape that is commonly seen in industrial grippers (Fig.~\ref{fig:gripperlook}a).
We fix the spacing between the two palms to $8\,$cm unless otherwise mentioned.
The four pneumatic actuators are connected to a single pressure supply, and a pressure of $120\,$kPa is enough to cause large bending of the palms and close their gap (Fig.~\ref{fig:gripperlook}b).
The two variable-stiffness units are connected to a single vacuum pump.

\subsection{Grippers with Different Jamming Media and Shape}
In addition to the above gripper, which we refer as F1 for convenience, we fabricate three other grippers that differ in the jamming media or gripper shape from the above design.
To explore the effect of plastic sheet in gripper performance, we make a gripper (F2) that does not have the plastic sheet liner.
We also investigate the difference between our fabric jamming-based gripper and layer jamming-based gripper by making a gripper (L1) in which the fabrics are replaced by a stack of papers with equal weight ($22\,$g) and size ($10\times8\,$cm), resulting in 32 pieces of papers.
The paper stack is placed in a volume with smaller thickness ($6\,$mm) than that of the fabrics, and the resulting palm thickness is also smaller.
We also explore the effect of gripper palm shape on gripper conformability by using a trapezoidal shape palm (Fig.~\ref{fig:fabrication}f right) and a gripper (T1).
The base lengths of the trapezoid are $5\,$cm and $11\,$cm, respectively.

\section{Characterization of the gripper}\label{sec:characterization}
\subsection{Object Gripping Performance}\label{sec:objectgrip}
We demonstrate the gripper's versatility by successfully picking up $12$ common and representative objects using the F1 gripper, shown in Fig.~\ref{fig:objectGripping} (also see supplementary video 1).
The gripper is mounted on a universal testing machine (Mark-10 F305).
We apply a positive pressure of $120\,$kPa to the pneumatic actuators so that the palms can conform and grip the object.
When necessary, the variable-stiffness structure is vacuumed to $-90\,$kPa to activate jamming. 
We then lift the gripper at a constant speed of $0.5\,$m/min.
During the test, the gripper can successfully pick up a grape berry ($10\,$g, Fig.~\ref{fig:objectGripping}a), a bok choy ($78\,$g, Fig.~\ref{fig:objectGripping}b), an apple ($192\,$g, Fig.~\ref{fig:objectGripping}c), a pitaya ($454\,$g, Fig.~\ref{fig:objectGripping}d), a bag of vegetable fries ($55\,$g, Fig.~\ref{fig:objectGripping}e), a stapler ($35\,$g, Fig.~\ref{fig:objectGripping}f), and a hand cream ($124\,$g, Fig.~\ref{fig:objectGripping}g), demonstrating its great capacity in gripping common items of a wide range of size and weight, as well as soft and deformable food packages.
Note that the grasping is accomplished by the variable-stiffness structures which is in direct contact with the items, not the soft pneumatic actuators.
For small lightweight objects such as the grape berry and the stapler, friction is sufficient to grip them.
For heavy objects such as the pitaya, jamming is required to stiffen the palm so that a strong geometric constraint can be imposed to enclose the object and passively provide gripping force.
In the supplementary video we show the lifting process of each object.

We also test delicate objects such as a badminton shuttlecock ($4\,$g, Fig.~\ref{fig:objectGripping}h) and a ping pong ball ($3\,$g, Fig.~\ref{fig:objectGripping}i).
The gripper can pick up these objects without causing feather bending or indentation on the ball.
In addition, our gripper design can be easily adapted to grip slender or slippery objects such as a chocolate bar ($54\,$g, Fig.~\ref{fig:objectGripping}j), a bottle ($404\,$g, Fig.~\ref{fig:objectGripping}k) and a wet hydrogel sphere ($83\,$g, Fig.~\ref{fig:objectGripping}l), by decreasing the palm spacing to $6\,$ or $7\,$cm.

\begin{figure}[!htbp]
    \centering
    \framebox{\parbox{3.2in}{\includegraphics[width=3.2in]{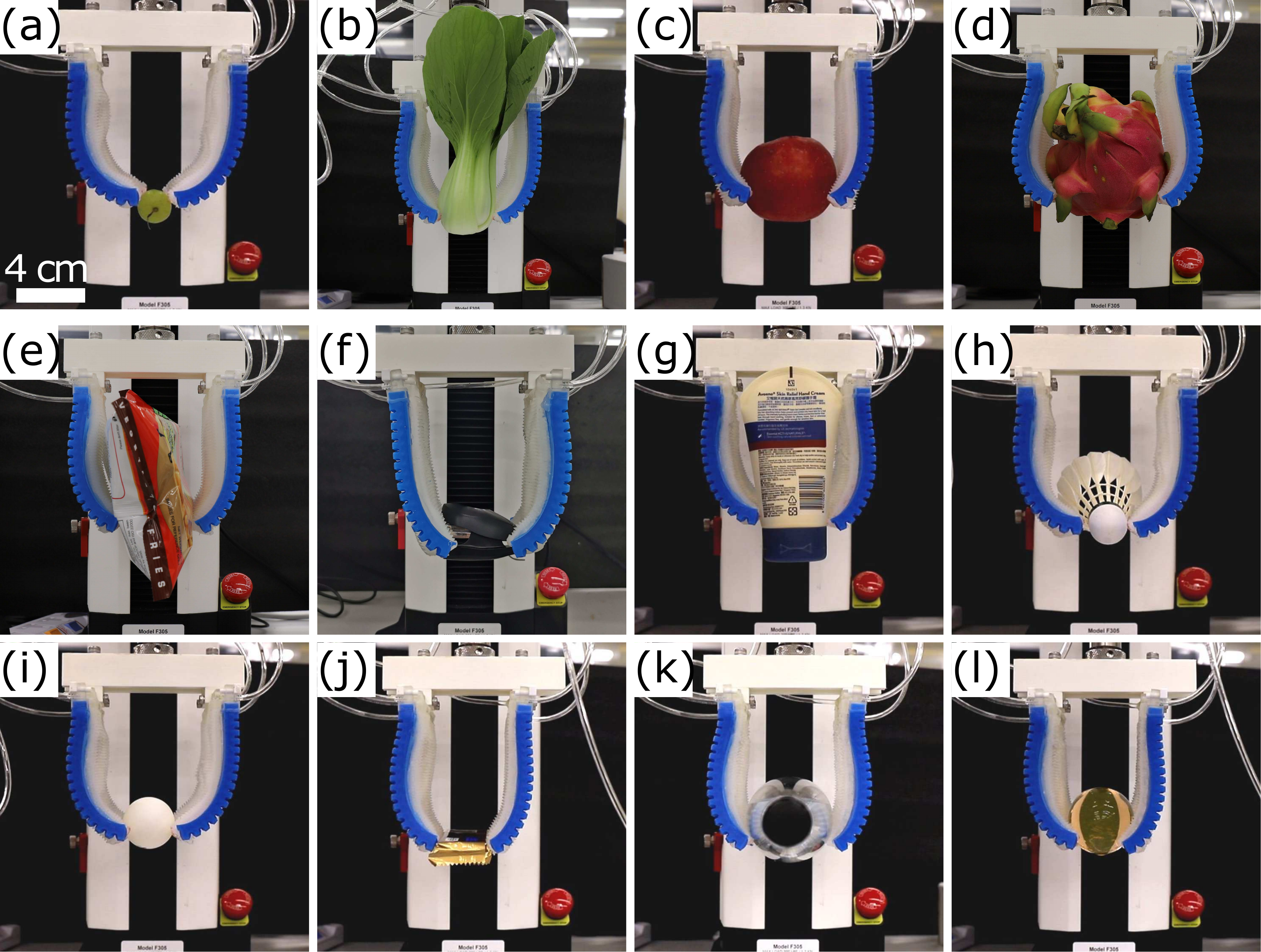}}}
    \caption{Object gripping test shows that gripper F1 can pickup a variety of objects. (a) grape berry ($10\,$g); (b) bok choy ($78\,$g); (c) apple ($192\,$g); (d) pitaya ($454\,$g, vacuum on); (e) vegetable fries ($55\,$g); (f) stapler ($35\,$g); (g) hand cream ($124\,$g); (h) badminton shuttlecock ($4\,$g); (i) ping pong ball ($3\,$g); (j) chocolate bar ($54\,$g); (k) bottle ($404\,$g, vacuum on); (l) wet hydrogel sphere ($83\,$g, vacuum on).}
    \label{fig:objectGripping}
\end{figure}

\subsection{Maximum Gripping Force}\label{sec-maxgrip}
\begin{figure}[!htbp]
    \centering
    \framebox{\parbox{3.2in}{\includegraphics[width=3.2in]{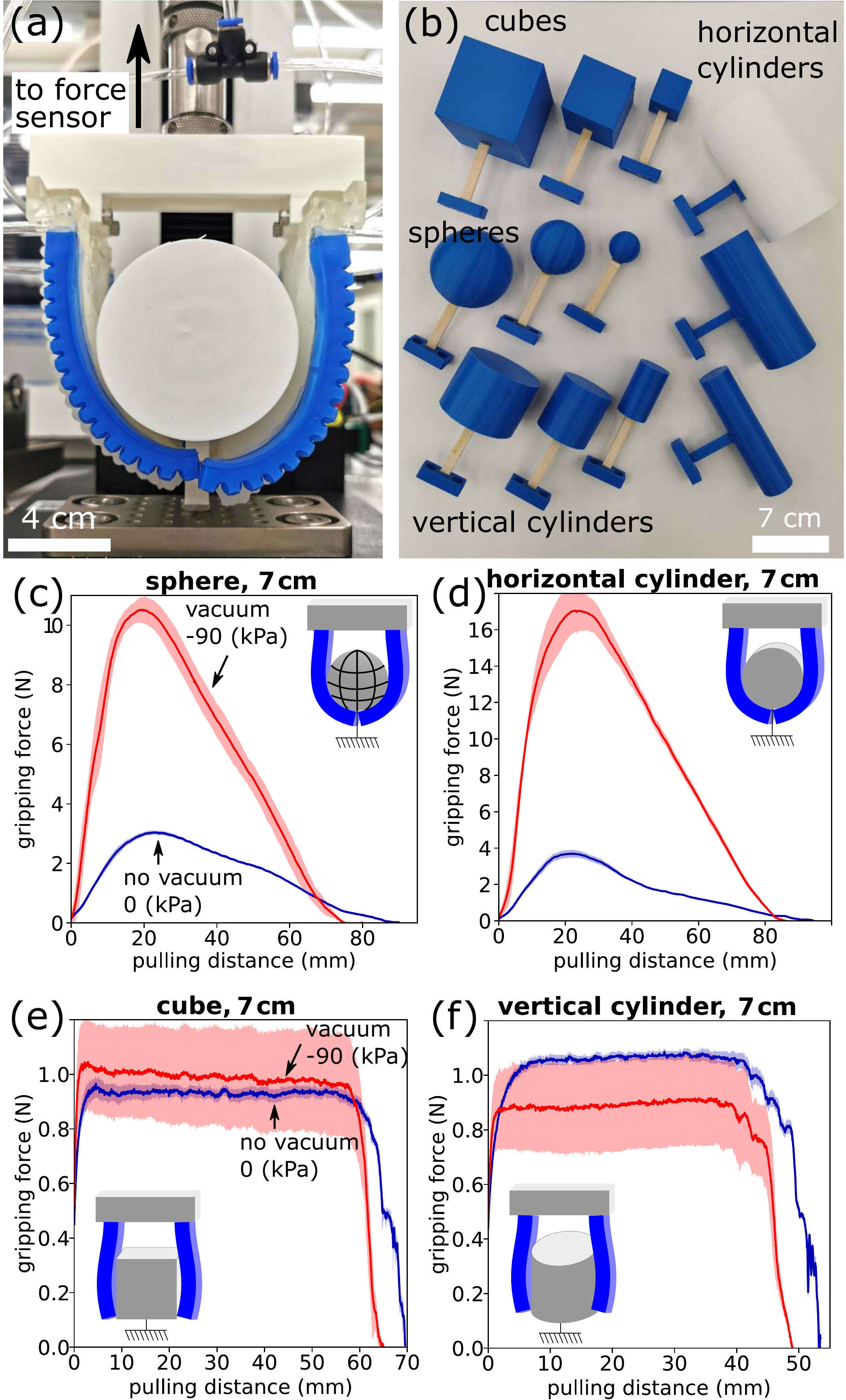}}}
    \caption{Gripping force test. (a) measurement setup; (b) the $12$ shapes used in the gripping force test; (c)-(f) gripping force vs.\ displacement curves of spheres, horizontal cylinder, cube and vertical cylinder, respectively. The diameter or length of the objects are all $7\,$cm. The red/dark blue curves are obtained with jammed (vacuum $-90\,$kPa)/unjammed palms. Each curve is an average of $5$ independent repeated runs. The insets illustrate gripping postures.}
    \label{fig:maxForce}
\end{figure}

A universal testing machine (Mark-10 F305) is used to pull the gripper off a fixed object, as illustrated in Fig.~\ref{fig:maxForce}a, and the pulling force is measured as a function of displacement.
The gripper is mounted onto the moving force sensor of the testing machine, and the weight of the gripper is offset so that we can measure the pulling or pushing force the gripper experienced.
We create $12$ testing objects with different shapes and sizes, shown in Fig.~\ref{fig:maxForce}b.
The shapes are spheres, cubes, and cylinders that orient either horizontally or vertically.
The diameters of the spheres and cylinders, and the lengths of the cubes are $7\,$cm, $5\,$cm, and $3\,$cm for each object.
For a single run of the experiment, the gripper is first lowered to a fixed position, then the soft actuators are pressurized to $120\,$kPa to grasp the object.
This position is chosen so that for spheres and horizontal cylinders, the gripper palms completely enclose the object; for cubes and vertical cylinders, the palms only cover up to the edge of the object (illustrated in the inset of Fig.~\ref{fig:maxForce}c-f).
The vacuum level in the variable-stiffness structures is controlled at either $0\,$ or $-90\,$kPa, so that we can measure the effect of jamming on gripping force.
We then pull the gripper at a constant speed of $2\,$mm/s until it detaches from the object.
The same test is repeated for $5$ times for each object.

We observe two qualitatively different responses: 
for spheres (Fig.~\ref{fig:maxForce}c) and horizontal cylinders (Fig.~\ref{fig:maxForce}d), the gripping force first increases with the pulling distance, reaching a maximum value, then decreases to zero when the gripper loses contact with the object.
For cubes (Fig.~\ref{fig:maxForce}e) and vertical cylinders (Fig.~\ref{fig:maxForce}f), the gripping force is smaller than those for spheres and cubes, and remains constant throughout the pulling process.
We identify two main mechanisms contributing to the maximum gripping force: geometric interlocking and frictional contacts.
For cubes and vertical cylinders, frictional contacts solely contribute to the gripping force.
For spheres and horizontal cylinders, the interlocking effect dominates the gripping force.
Additional effort is needed to break the geometric constraints by the bent palms, hence the gripping forces are larger than cubes and vertical cylinders.
Upon jamming, the increased palm stiffness results in stronger constraints and hence significantly larger gripping forces are measured when gripping spheres and horizontal cylinders.
The maximum gripping force increases $4.6$ times for the $7\,$cm sphere and $3.5$ times
for the $7\,$cm  horizontal cylinder.

While for cubes and vertical cylinders, the gripping force changes slightly before and after jamming.
Because the gripping force is mainly due to frictional contact in these cases, the result suggests that the pinch force, i.e. the force that the palms actively pinch the object, remain almost unchanged when jammed.
This assumption is justified by our pinch force measurement in Sec.~\ref{sec:block}.
The vacuum-induced jamming sometimes lead to additional small deformation of the palm, so that the palm end comes closer or leave further from the object.
This shape change by jamming can cause slight increase or decrease of the contact force, as measured by our pinch force test in Fig.~\ref{fig:blockingforce}c-e.
We attribute this effect to imperfections during manufacturing process.
This effect is less prominent for gripping spheres and horizontal cylinders, because the gripper forms a locking mechanism with these objects.

\begin{figure}[!htbp]
    \centering
    \framebox{\parbox{3.2in}{\includegraphics[width=3.2in]{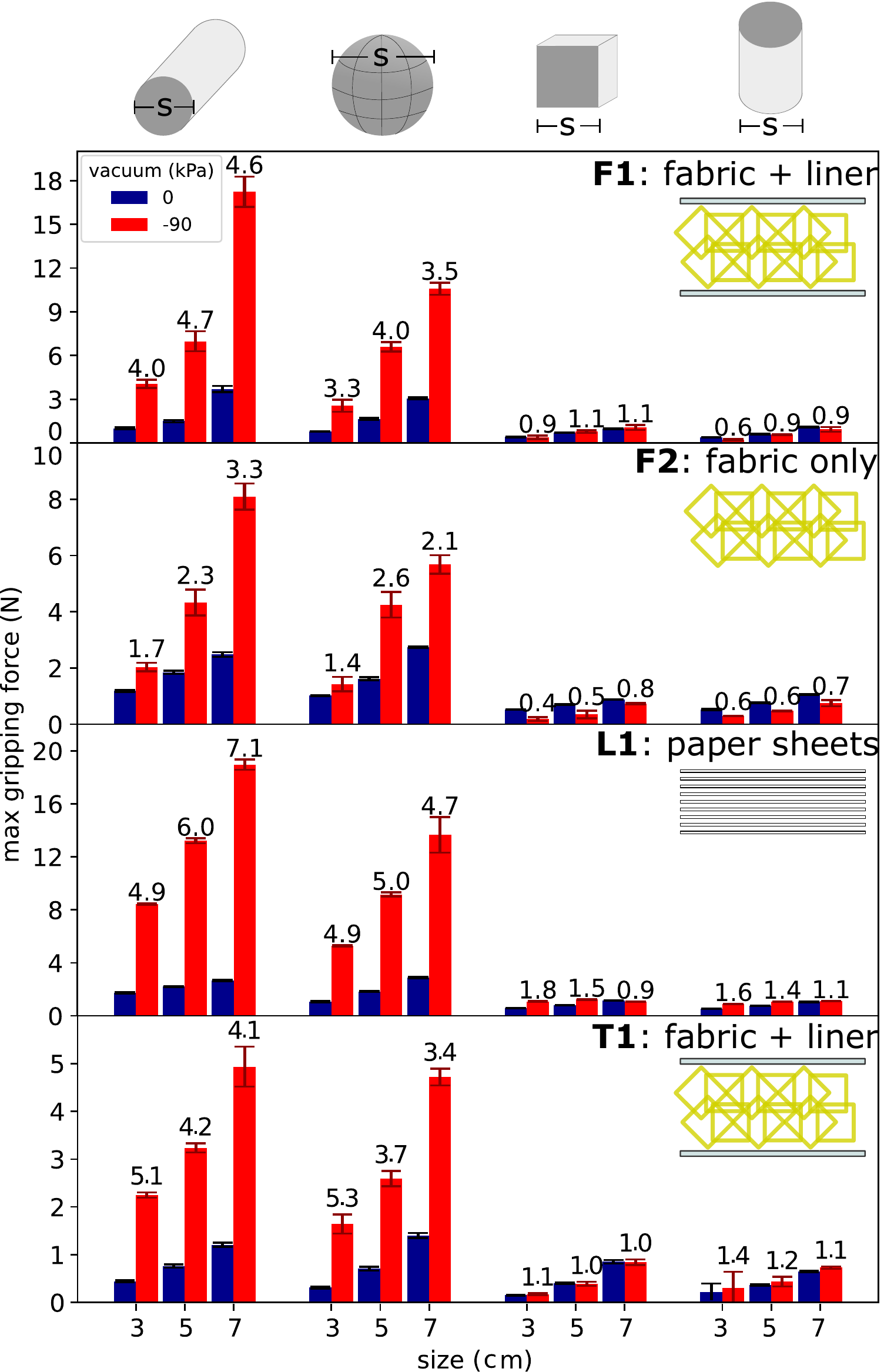}}}
    \caption{The averaged maximum gripping force $\langle F_{\rm max} \rangle$ is plotted for different object shapes and sizes. The object shapes are illustrated in the top row. Each of the four sub-figures shows $\langle F_{\rm max} \rangle$ for different gripper designs, which are F1, F2, L1 and T1 from top to bottom. Dark blue/red color represents $\langle F_{\rm max} \rangle$ measured in unjammed/jammed ( $-90\,$kPa) state. The ratio of $\langle F_{\rm max} \rangle$ between jammed and unjammed case is marked on top of each bar.
    The insets illustrate each gripper composition.
    }
    \label{fig:shapeDependence}
\end{figure}

The gripping force responses are qualitatively the same for the F2, L1 and T1 grippers, and we summarize the results by comparing the averaged maximum gripping force $\langle F_{\rm max} \rangle$ across different gripper designs, object shapes and sizes in bar charts in Fig.~\ref{fig:shapeDependence}.
We find that $\langle F_{\rm max} \rangle$ increases with object size, with a stronger dependence for spheres and horizontal cylinders and a weaker dependence for cubes and vertical cylinders.
We also find that upon jamming, $\langle F_{\rm max} \rangle$ noticeably increases for spheres and horizontal cylinders, but not for cubes and vertical cylinders, due to slight volume decrease of the palms upon jamming as we discussed before.
The ratio of $\langle F_{\rm max} \rangle$ between jammed and unjammed configurations are shown on top of each bar.
The highest ratio is $7.1$ achieved by the L1 gripper on horizontal cylinder of diameter $7\,$cm.
Jamming effectively improves the gripping strength of L1 gripper, and the ratio ranges from $4.7$ to $7.1$.
For F1, F2 and T1 gripper, the ratio ranges from 3.3 to 4.7, 1.4 to 3.3, and 3.4 to 5.3, respectively.
Notably, L1 gripper has the largest $\langle F_{\rm max} \rangle$ on most shapes, achieving an $\langle F_{\rm max} \rangle$ of $19\,$N on $7\,$cm horizontal cylinder.
For the same object, $\langle F_{\rm max} \rangle$ is $17\,$N for F1 gripper, $8\,$N for F2 gripper, and $4.8\,$N for T1 gripper.
When the variable-stiffness structure is unjammed, $\langle F_{\rm max} \rangle$ is similar for different gripper designs.

\subsection{Pinch Force Test}\label{sec:block}
\begin{figure}[!htbp]
    \centering
    \framebox{\parbox{3.2in}{\includegraphics[width=3.2in]{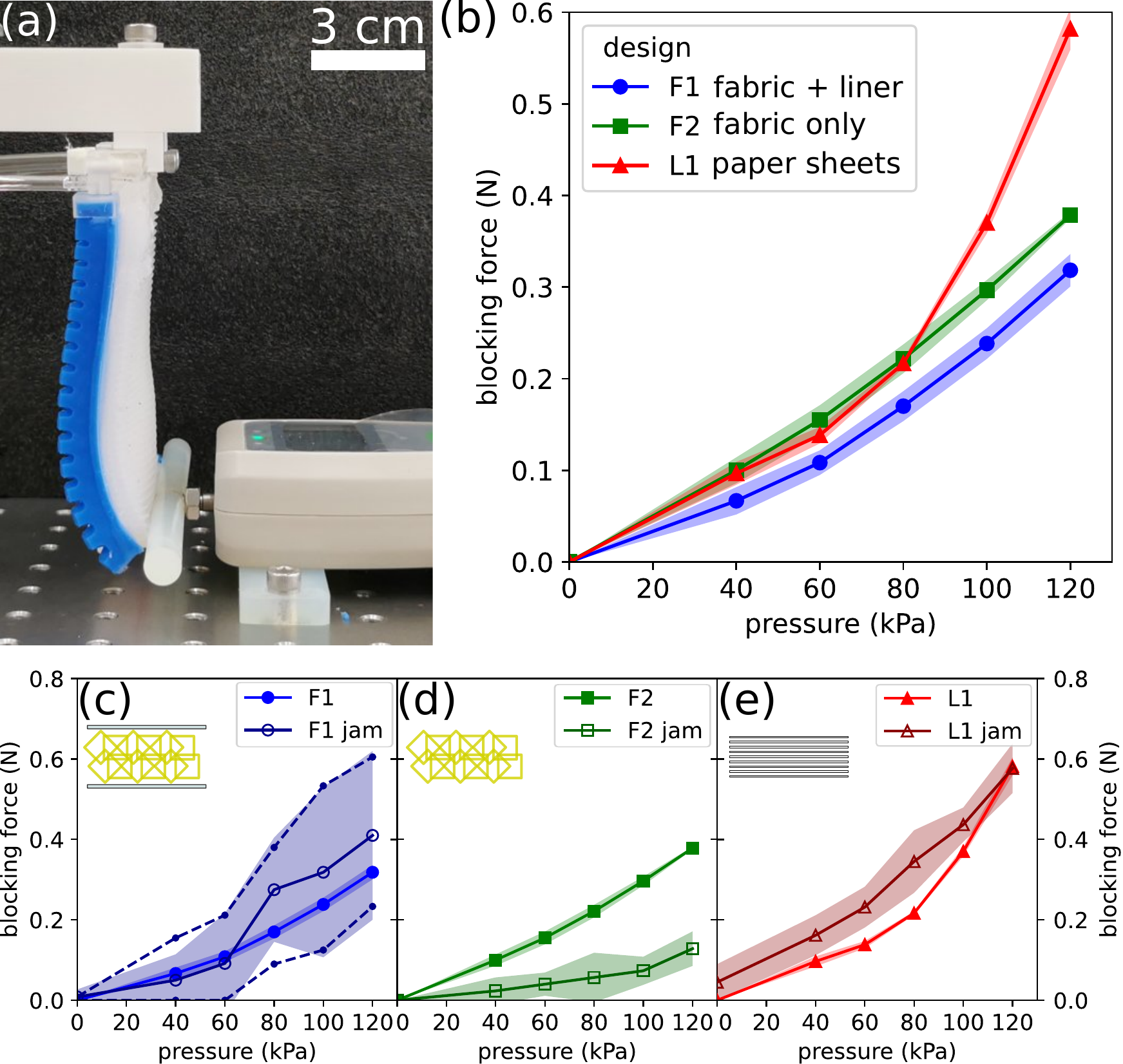}}}
    \caption{Pinch force test. (a) the setup: a palm is pressurized to $120\,$kPa with its end pressing a rod connected to a force sensor; (b) pinch force measurement is plotted as a function of actuator pressure for different gripper designs, F1 (blue circles), F2 (green squares), and L1 (red triangles). Each curve is an average of $6$ independent measurements from the two gripper palms in the gripper;
(c-e) pinch force measurements for the unjammed palms (solid symbols) and jammed palms at $-90\,$kPa (empty symbols) for the three gripper design.
The dashed dotted lines in (c) are average pinch force for the individual palm in the jammed state.
The insets illustrate each gripper composition.}
    \label{fig:blockingforce}
\end{figure}

To estimate the gripper's pinch force, we perform pinch force tests on the gripper palm, shown in Fig.~\ref{fig:blockingforce}a.
The pinch force as a function of actuator pressure for the three palm designs are shown in Fig.~\ref{fig:blockingforce}b, and the effect of jamming is shown in Fig.~\ref{fig:blockingforce}c, d and e.
All pinch forces increase as a function of pressure, and the results of both palms are averaged.
When the palm is unjammed, the L1 gripper has the maximum pinch force of $0.58\,$N at $120\,$kPa.
F2 gripper has a pinch force  of $0.38\,$N, higher than $0.32\,$N of F1 gripper at $120\,$kPa.
Since all palms use the same soft pneumatic actuators, we think the difference mainly comes from the bending resistance of the variable-stiffness structure when it is bent in one direction.
Comparing F1 and F2, their difference is mainly due to the insertion of the PET liner that has a larger constraint to bending, resulting in lower pinch force.
When the palms are first actuated and then jammed at $-90\,$kPa, the pinch force changes slightly for the F1 and L1 grippers (Fig.~\ref{fig:blockingforce}c-e), and becomes small for the F2 gripper.
This shows that jamming has a small effect on the pinch force.
Note that even though pinch force decreases for the F2 gripper when jammed, the gripping forces for spheres and horizontal cylinders increase up to 2.6 and 3.3 times, respectively (Fig.~\ref{fig:shapeDependence}), which is due to the geometric interlocking effect when jammed.
When only the frictional contact mechanism exist, such as gripping cubes and vertical cylinders, the decrease in pinch force for the F2 gripper results in a consistently smaller gripping force as shown in Fig.~\ref{fig:shapeDependence}.

We also calculate the ratio between maximum gripping force $\langle F_{\rm max} \rangle$ ($7\,$cm horizontal cylinder in the jammed state) and pinch force at $120\,$kPa when jammed, and they are about $42\,$, $62\,$, and $33\,$ for the F1, F2 and L1 grippers, all well surpassing current state-of-the-art in jamming-based variable-stiffness finger or membrane grippers (Fig.\ref{fig:gripperlook}e).
The large ratios show that our gripper actively applies a very small force for picking up delicate objects, and can passively provide large force due to the jamming fabric-induced geometric interlocking mechanism~\cite{brown_2010_universal}.

Fig.~\ref{fig:blockingforce}c also shows a large fluctuation in the pinch force when jammed, as indicated by the individual palm pinch force measurement (the dash-dotted lines) where one palm has larger pinch force than the unjammed case and the other has lower force.
This is caused by a noticeable shape change in the palm during vacuum jamming.
We believe that this is a combined effect due to the plastic deformation and residue stresses between the plastic liner and the fabric, the variation in random packings of the fabrics, and the imperfections introduced during fabrication such as misalignment of the actuator and the silicone membrane.
Nevertheless, the variation in pinch forces has a very small effect on the gripper performance since our gripper mainly relies on the geometric interlocking mechanism for holding rather than the pinch force.

\subsection{Palm Conformability Test}
\begin{figure}[!htbp]
    \centering
    \framebox{\parbox{3.2in}{\includegraphics[width=3.2in]{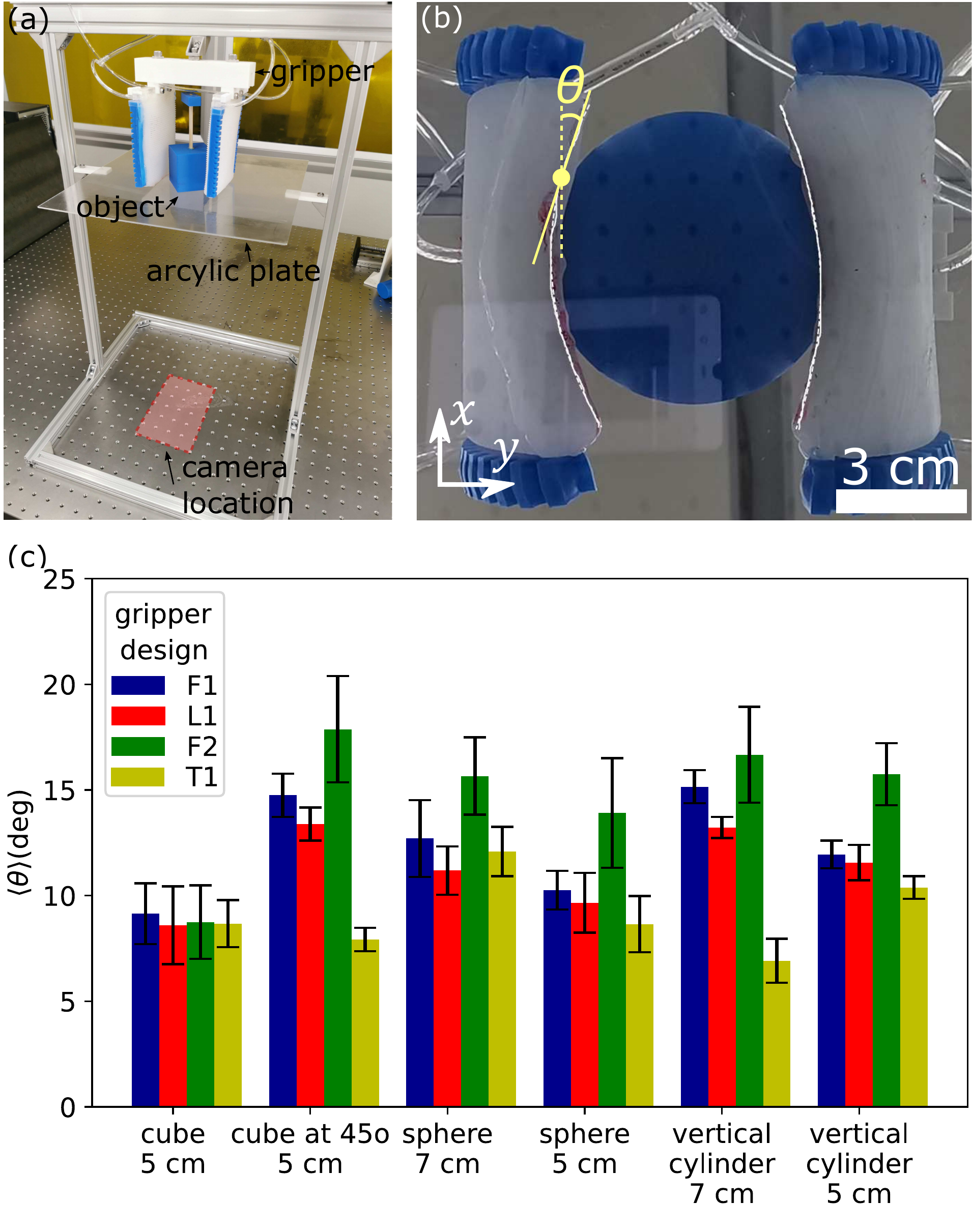}}}
    \caption{Gripper conformability test. (a) test setup; (b) an image of the F1 gripper holding a $7\,$cm sphere. White solid lines are the traced palm bottom edges. White dashed lines are parabola fitting to the edges. $\theta$ is the slope angle between the edge and $x$ axis. The coordinate system used in data processing is in the lower-left corner; (c) bar chart of average angle $\langle\theta\rangle$ verses different objects for three gripper designs, F1 (darkblue), F2 (green), L1 (red), T1 (yellow).}
    \label{fig:conformability}
\end{figure}

Conformability generally refers to the ability of a flexible object to adapt its shape or structure to the environment.
Depending on the object or material, various quantification methods exist such as for textiles~\cite{sageman-furnas_2014_sphereprint}, plastic films~\cite{zhang_2020_time-dependent}, and stretchable electronics~\cite{dong_2017_theoretical}.
We adopt a simple method to characterize and compare the conformability of the palms, by using the palm's bottom edge bending deformation.
The measurement setup is shown in Fig.~\ref{fig:conformability}a.
The gripper is hang above a transparent acrylic plate, which allows the palms and the object to be captured by a camera viewing from below.
The gripper is able to grasp an object on the plate, and we track the deformation of the bottom edge of the palms when the gripper conforms to object surface.
A typical captured image is shown in Fig.~\ref{fig:conformability}b, in which a $7\,$cm sphere is grasped by the F1 gripper.
To reduce image distortion, the image data is calibrated via MATLAB \texttt{cameraCalibrator} and a checkerboard pattern.
We then extract the coordinates of the two edges from each image (illustrated in Fig.~\ref{fig:conformability}b), which are used to compute an average angle $\langle\theta\rangle$ (for each edge) given by:
\begin{equation}
\langle \theta \rangle = \frac{\int_{x_l}^{x_u}\theta (x) dx}{x_u - x_l}
\end{equation}
where $\theta(x)$ is the unsigned slope angle between the curve and $x$ axis (Fig.~\ref{fig:conformability}b).
$x_u$ and $x_l$ are the upper and lower end of the curve in the x direction.
$\theta(x)$ is averaged at all points along the curve.
To facilitate the averaging, we fit the curve to a parabola $y=a x^2 + b x + c$, and the fitting coefficients $a$ and $b$ are used to compute the average angle $\langle\theta\rangle$.
The value of $\langle\theta\rangle$ reflects how much the bottom edge of the palm is bent when conforming to object surface under the actuation of the two actuators on the side.
The greater $\langle\theta\rangle$, the larger conformability it has.

We use 6 objects in the test, including spheres of diameters $7\,$cm and $5\,$cm, vertically placed cylinders of diameters $7\,$cm and $5\,$cm, and cubes of length $5\,$cm with two orientations, either parallel or at $45^\circ$ angle with respect to the palm surface (Fig.~\ref{fig:conformability}b).
For each object, we pressurize the actuators to $120\,$kPa and repeat three time independently.
The resulting $\langle\theta\rangle$ for different objects are shown in Fig.~\ref{fig:conformability}c.
We find that for most objects, F2 gripper has the highest $\langle\theta\rangle$.
F1 and L1 have smaller $\langle\theta\rangle$ than F2, with F1 being slightly better than L1.
For the $5\,$cm cube, all three grippers have similar $\langle\theta\rangle$.
It is likely due to the flat contact between the gripper palm and the object surface, so that the palms don't bend significantly.
Our results suggest that structured fabrics jamming-based grippers have a better conformability than layer jamming-based gripper.
This is expected because the discrete fabrics requires little elastic energy to deform its shape when adapting surfaces, while for elastic layers the energy is non-negligible~\cite{zhang_2020_time-dependent}.
T1 gripper, however, has the lowest $\langle\theta\rangle$ over all shape tested.
This is likely due to the arrangement of actuators that lead to less palm bending.

\section{Discussion and conclusion}\label{sec:conclusion}
We fabricate and test a novel soft variable-stiffness gripper based on jamming of structured fabrics, and demonstrate its good capacity in picking up various objects.
The gripper consists of two soft palms driven by pneumatic actuators (Fig.~\ref{fig:gripperlook}).
The palms can significantly increase their stiffness  when jammed, enabling the gripper to form a rigid enclosure on the object for successful gripping.
We are able to reach a maximum gripping force of $17\,$N for the fabric jamming gripper (F1) with the addition of plastic sheet liners, and $19\,$N for the layer jamming gripper (L1).
Both grippers can pick up soft and slippery objects such as a wet hydrogel spheres with its jamming capacity.
Our gripper is also conformable due to the discrete nature of the fabrics, and we conduct quantitative characterization of gripping conformability.
We further characterize our gripper's pinch force by measuring the pinch force of a single palm.
The pinch force is in the range from $0.2\,$N to $0.6\,$N, and a maximum gripping-to-pinch force ratio of 42 can be achieved by the fabric jamming gripper (F1) when jammed.
The high gripping-to-pinch force ratio and good conformability in our palm design improve over current universal gripper design and jamming-based anthropomorphic soft grippers, where the former has good conformability but requires large pushing force, while the latter can handle delicate objects but lacks conformability and stiffness.
The large gripping-to-pinch force ratio also makes it ideal for handling delicate items.
Our gripper design is also a step towards utilizing robotic surface~\cite{shah_2021_shape} for gripping tasks.
We anticipate that our gripper design has a wide range of applications in packaging, logistics, food and agriculture tasks.

However, our current designs have limitations in bending deformation because it is only driven by two bending actuators, and this simplicity may reduce its conformability.
Future works can be done to optimize the actuator placements for higher gripper conformability and better gripping performance.
The gripper with the max gripping force (L1, Fig.~\ref{fig:maxForce}) has the lowest conformability (Fig.~\ref{fig:conformability}), which might suggest a negative correlation between conformability and gripping force in our current design, since more compliant materials typically reach lower stiffness under jamming.
Future works will focus on further improving the stiffness variation of fabric jamming by designing different fabric structures and fabric-liner-membrane compositions, and increasing conformability by using more effective actuation mechanisms such as optimized bending actuator placement, stimulus-responsive materials and more powerful pneumatic actuators driven by either positive or negative pressures, while the current pneumatic control system can be further compacted by exploring vacuum or electric-powered actuators.

\balance

\bibliographystyle{IEEEtran}
\bibliography{IEEEabrv,mybibfile}

\end{document}